# OPTIMIZED OBJECT TRACKING TECHNIQUE USING KALMAN FILTER

**Liana Ellen Taylor [a,*], Midriem Mirdanies [b], Roni Permana Saputra [b]**

[a]School of Engineering and Information Technology - University of New South Wales (UNSW)
Canberra, ACT 2600, Australia
[b]Research Center for Electrical Power and Mechatronics, Indonesian Institute of Sciences (LIPI)
Komplek LIPI Bandung, Jl. Sangkuriang, Gd. 20. Lt. 2, Bandung 40135, Indonesia

**Abstract**

This paper focused on the design of an optimized object tracking technique which would minimize the processing time required in the object detection process while maintaining accuracy in detecting the desired moving object in a cluttered scene. A Kalman filter based cropped image is used for the image detection process as the processing time is significantly less to detect the object when a search window is used that is smaller than the entire video frame. This technique was tested with various sizes of the window in the cropping process. MATLAB® was used to design and test the proposed method. This paper found that using a cropped image with 2.16 multiplied by the largest dimension of the object resulted in significantly faster processing time while still providing a high success rate of detection and a detected center of the object that was reasonably close to the actual center.

Keywords: Kalman filter; object tracking; object detection; cropping; color segmentation.

## I. INTRODUCTION

In the recent decades, the use of computer vision in many applications has been significantly increased from manufacturing application into military applications [1-3]. A major aspect of computer vision applications is vision based object tracking. Tracking objects in the real time environment is not a trivial task and has been a popular research topic in the computer vision field. A robust vision tracking system allows for many applications such as traffic monitoring, autonomous system guidance, surveillance systems, human computer interaction, vehicle navigation, and automated weapons systems [2, 4, 5].

Various algorithms and methods have been developed for object detection based on the object's color, feature points or other. Methods based on feature point detection have been carried out by Bing, using Speeded Up Robust Feature (SURF) tracker [6] and Fazli, using Scale Invariant transform Feature (SIFT) tracker [7]. In previous research, modification of SIFT and SURF methods has been performed to detect multiple objects at once [2]. These methods are accurate but require higher processing time. Methods based on color have been researched by Comaniciu using Mean-shift [8] and Wang using color distribution [9]. These methods require less processing time but the accuracy is limited.

For object detection, a fundamental feature is the robustness and resistance to changes in the scene image, such as different lighting conditions and blurriness of the object due to rapid movement [5]. Another feature that is crucial to object tracking implementation, especially for real-time applications, is computational time. To be feasibly implemented in a real-time process, such as for robotic sensor application, the object detection process has to be implemented efficiently so that is may run fast. Even though the detection process is relatively fast, the system must still maintain its performance in terms of success rate, accuracy, and repeatability of the detection process.

In this research, an algorithm of fast and accurate visual object tracking is presented. The algorithm is designed to improve object-tracking process by minimizing the processing time required while maintaining accuracy in detecting the desired object in a cluttered scene. In this approach, a combination of Kalman filter estimation technique and color segmentation based object recognition is used. Parameters of the process were varied to determine the relationship between each parameter and the performances of the object tracking process. Based on the experimental results, the optimum set of parameters can then be determined to obtain the desired performance of the object tracking process. The research was undertaken at the Mechatronics Lab. Research Center for Electrical Power and Mechatronics - Indonesian Institute of Sciences (LIPI).

---

* Corresponding Author. Tel: +61-439621755
E-mail: lianat36@gmail.com

## II. METHODOLOGY

### A. Color Based Object Detection From The Video

The method used for object recognition in this paper was detection of the object based on color segmentation. The advantage of using color segmentation detection was that it could support blurred images of the object. Blurred images of the object are common when an object is moving rapidly in a video, which was needed to test the Kalman filter. The objects used in this experiment have red as their dominated color. Each object was detected from a frame acquired from a video in RGB format. The red layer of the video frame was extracted and a gray scale of the frame was subtracted from the red frame. Noise was filtered out by a median filter. A binary image was then formed with a red threshold of 0.25 to obtain a binary image containing 'binary large objects' (blob).

Since the objects detected in this experiment have red dominate color, the largest blob resulting from the described process would be the object. Based on this largest detected blob, the bounding box then can be obtained and used to acquire the center location of the object. The step-by-step object detection process in this method can be seen in the Figure 1. This method can also be used to detect and track another object with different color by adjusting the threshold on color segmentation process.

### B. Kalman Filter Based Estimation

Kalman filter is one of the most common approaches for estimating linear state that is assumed to have a Gaussian distribution [10]. This technique is published first by R.E. Kalman, in 1960 as a recursive solution for linear filtering in discrete data [11]. The Kalman filter is an efficient recursive process that provides estimation of the states at the past, present, and future time domain. It also estimates the state while the model and the nature of the system is not known precisely [12].

*1) Basic Kalman Filter Operation*

In general, the operation of discrete time Kalman filtering has the form that can be seen in Figure 2. There are two main operations, the prediction step and correction step. This operation form is similar to feedback control where at filter estimates the state in the prediction step, and then obtains feedback in the correction step based on the measurement result [12].

The prediction step in this process is responsible for predicting the states to obtain *a priori* estimation of the next step based on the projection of the current state. Meanwhile, the correction step in this process will improve the *a priori* estimation resulting in the prediction step, based on obtained measurement results and obtains *a posteriori* estimation. For discrete time systems, these two operations are expressed using the notations as shows in Figure 2.

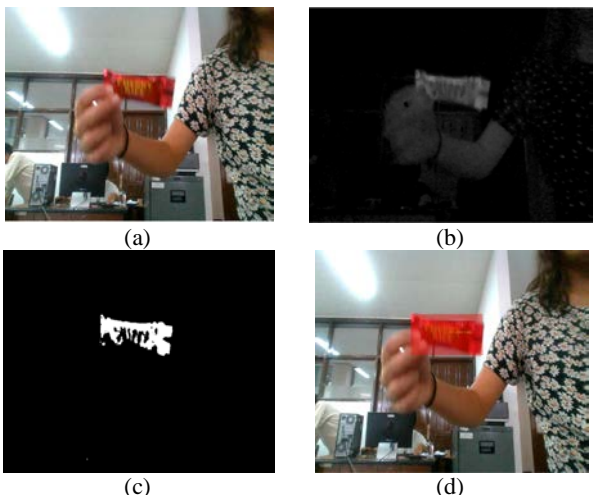

(a)   (b)
(c)   (d)

Figure 1. Step-by-step object detection based on color segmentation (a) Original RGB image; (b) Gray scale of the image subtracted from the red layer; (c) Binary image with pixels above the red threshold of 0.25 converted to white; (d) After blob analysis of image, a red box is drawn around the box with the center indicated

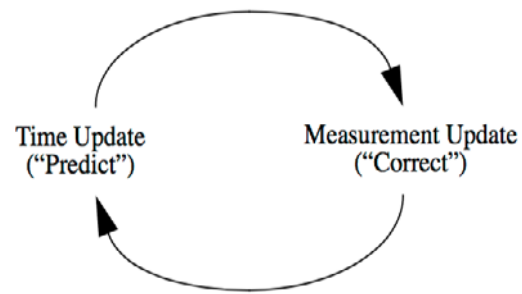

Figure 2. Basic operation of the Kalman filter

- $\overline{X}(k+1|k)$ → state estimation at time k+1 predicted based on the parameter at time k
- $P(k+1|k)$ → covariance of the estimated state at time k+1.
- $\overline{X}(k|k)$ → state estimation at time k updated based on the observation process or measurement at time k.
- $P(k|k)$ → covariance of the estimated state at time k.

*2) Prediction Step*

The mathematical expression for this '*time update*' step can be seen in the Eq. (1) and 2 [13]. In these equations, the future state ($\overline{X}(k+1|k)$) is predicted based on the process model **F** and the current state ($\overline{X}(k|k)$). The prediction covariance ($P(k+1|k)$) is also calculated based on the process model F and the uncertainty of the process model ($Q(k)$).

$$\overline{X}(k+1|k) = F\overline{X}(k|k) \quad (1)$$

$$P(k+1|k) = FP(k|k)F^T + Q(k) \quad (2)$$

*3) Correction Step*

The next operation in the Kalman filter is the correction step, also known as the '*measurement update*' step. In this step, the predicted state is corrected based on the difference between the real measured result and the expected measurement result ($\overline{z}_{k+1}$) from the measurement model (H). Equation 3 shows the mathematical expression of the measurement model [13].

$$\overline{z}_{k+1} = H\overline{X}_{k+1} \quad (3)$$

Once the real measured result is obtained, the difference ($v$) between the real measurement and the measurement model is calculated in Eq. (4). The Kalman gain, K, is then calculated to update the estimated state $\overline{X}_{k+1|k}$ to become $\overline{X}_{k+1|k+1}$. The complete calculation of this correction step is expressed in Eq. (5) to Eq. (7) [13].

$$v_{k+1} = z_{k+1} - \overline{z}_{k+1} \quad (4)$$

$$K_{k+1} = P_{k+1|k}H^T(HP_{k+1|k}H^T + R_{k+1|k})^{-1} \quad (5)$$

$$\overline{X}_{k+1|k+1} = \overline{X}_{k+1|k} + K_{k+1} \cdot v_{k+1} \quad (6)$$

$$P_{k+1|k+1} = (I - K_{k+1}H)P^-_{k+1|k} \quad (7)$$

*4) Kalman Filter for Object Tracking*

In object tracking implementation, the Kalman filter predicts the object's next position from the previous state information about the object. It then verifies the result of the prediction using the result of the object detection process in the next following step.

In this paper, the object was assumed to move with constant velocity. Based on this, the motion model for this object tracking can be expressed in Eq. (8), while Δx and Δy are constant.

$$x_{k+1} = x_k + \Delta x$$

$$y_{k+1} = y_k + \Delta y \quad (8)$$

From Eq.(8), the motion model can be expressed in the form in Eq. (9) from Eq. (1).

$$\begin{bmatrix} x_{k+1|k} \\ y_{k+1|k} \\ \Delta x_{k+1} \\ \Delta y_{k+1} \end{bmatrix} = \begin{bmatrix} 1 & 0 & 1 & 0 \\ 0 & 1 & 0 & 1 \\ 0 & 0 & 1 & 0 \\ 0 & 0 & 0 & 1 \end{bmatrix} \begin{bmatrix} x_{k|k} \\ y_{k|k} \\ \Delta x_k \\ \Delta y_k \end{bmatrix}$$

$$F = \begin{bmatrix} 1 & 0 & 1 & 0 \\ 0 & 1 & 0 & 1 \\ 0 & 0 & 1 & 0 \\ 0 & 0 & 0 & 1 \end{bmatrix} \quad (9)$$

Based on the predicted state, the expected measurement result can be calculated as expressed in Eq. (10).

$$\overline{z}_{x(k+1)} = x_{k+1}$$

$$\overline{z}_{y(k+1)} = y_{k+1} \quad (10)$$

From Eq. (10), the measurement model can also be expressed from Eq. (3) to become Eq. (11).

$$\begin{bmatrix} \overline{z}_x \\ \overline{z}_y \end{bmatrix} = \begin{bmatrix} 1 & 0 & 0 & 0 \\ 0 & 1 & 0 & 0 \end{bmatrix} \begin{bmatrix} x_{k+1|k} \\ y_{k+1|k} \\ \Delta x_{k+1} \\ \Delta y_{k+1} \end{bmatrix}$$

$$H = \begin{bmatrix} 1 & 0 & 0 & 0 \\ 0 & 1 & 0 & 0 \end{bmatrix} \quad (11)$$

Figure 3 illustrates a summary of the Kalman filter process for image tracking implementation.

**C. Proposed Object Tracking Method**

In this paper, a Kalman filter based cropped image method is proposed. Figure 4 outlines the process. In this method, object tracking was implemented such that the object was searched for in a window that was smaller than the entire scene frame. The advantage of this is that it reduces the number of pixels that must be processed and minimizes the likelihood of detecting an object other than the object of interest in a cluttered scene. In addition, this method is also expected to be more robust in scenarios where the object is obstructed from view for several frames.

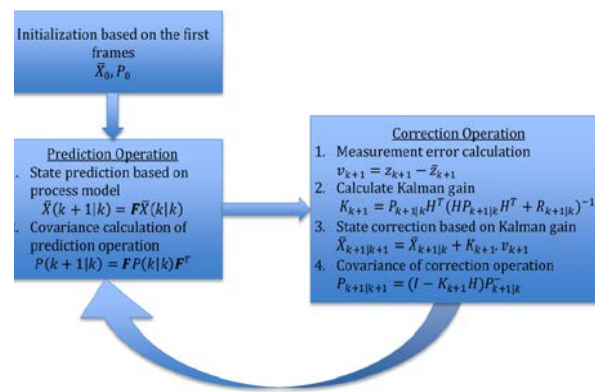

Figure 3. A summary of the Kalman filter operation

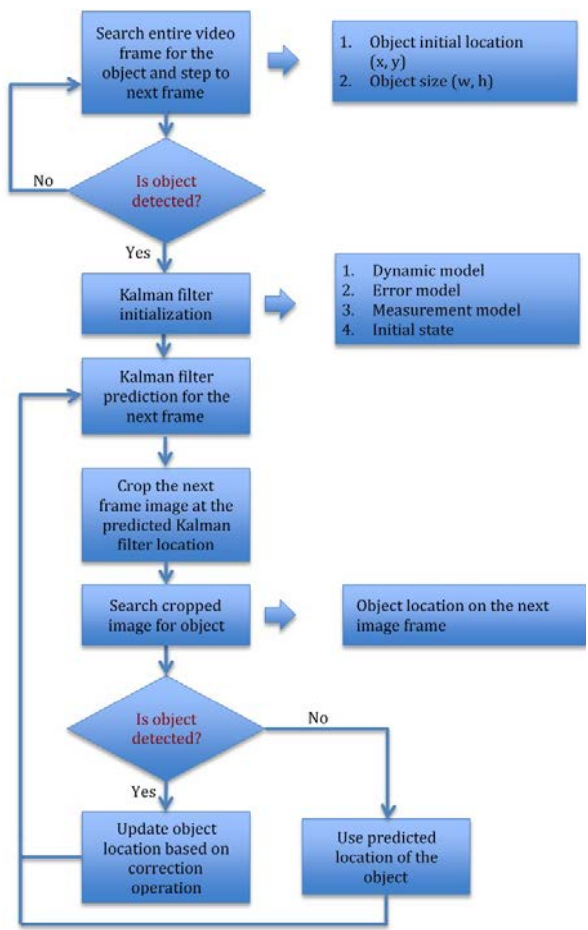

Figure 4. Method for implementing Kalman filter based cropped image

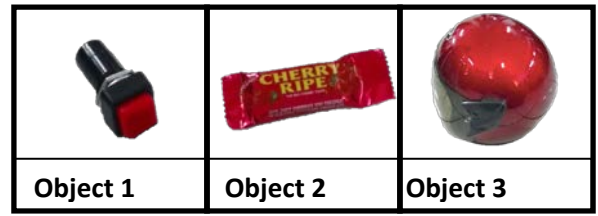

Figure 5. The various objects used for object tracking

There are four main steps in this method: initialization, Kalman filter prediction and image cropping, object detection on the cropped image and Kalman filter correction.

*1) Initialization*

The object is searched for in the entire image for each frame until it is detected. Once the object is detected, the Kalman filter is initialized with the object's detected location. The object's largest dimension is then used as a reference to determine the search window size for the entire video. The search window is set to be a width and height that is a specified multiple (referred to as the window multiple in this paper) of the object's largest dimension (i.e. width or height). In this initialization step, some parameters of the Kalman filter used in this operation are also configured such as: the motion model of the object, the error model of the prediction and measurement process, and the measurement model and the initial state.

*2) Kalman Filter Prediction and Image Cropping*

After the initial object location and size are detected and the Kalman filter is initialized, then the relative position for the object in the next frame is predicted by the Kalman filter. The scene image is then cropped to the search window located at the predicted location obtained as a result of the Kalman filter.

*3) Object Detection on The Cropped Image*

The object detection process in this method is performed on a smaller cropped image of the whole video frame located at the location obtained from the Kalman filter prediction. This approach is used to optimize the computational time of the detecting process while maintaining the accuracy.

*4) Kalman Filter Correction*

If the object is detected, the object's location is used to correct the state of the Kalman filter. This updated location is then used for prediction in the next iteration process. If the object is not detected, previous predicted location will be used for the next predicted location.

## III. MATERIALS

The objects selected for tracking in this experiment consisted of a push button, a Cherry Ripe bar and a motorbike helmet as shown in Figure 5. The objects were of various sizes (listed in Table 1) and shapes to determine if the optimal search window size was affected by the object size or shape. A video was shot of each object moving randomly in a scene and was used to provide an analysis of using a cropped image to search for the object in a given frame. The computer used in this experiment is a laptop with the following specifications: Intel(R) Processor 5Y10 CPU @ 1 GHz, 8 GB RAM and Microsoft Windows 1064 bit. The object tracking was coded in MATLAB 2014b and aimed to compare the accuracy and processing time of object tracking; both with and without image cropping of the scene image based on the object's previous location. The effect of the size of the cropped image window on accuracy and processing time was also observed.

Table 1.
Object sizes

| Object | Approximate pixel area (% of frame area) |
|---|---|
| Object 1 | 2.3 |
| Object 2 | 3.8 |
| Object 3 | 0.7 |

## IV. RESULTS AND DISCUSSION

Various search window sizes were used for testing. When the whole scene frame of the video was used for object detection, the object was detected 100% of the time for all objects. When the object was successfully found in a frame, its detected location was drawn to the video and displayed to the user as shown in (d) of Figure 1.

Each of the objects was successfully detected in all frames once the window multiple was greater than 1 (see Figure 6). Object 2 approaches a 100% success rate earlier than the other objects as it has a rectangular shape, and the longer width is chosen for the square search window. The processing time of the cropping method was compared to the processing time when searching the whole video frame as shown in Table 2. Object tracking was repeated 5 times for each window size for each object and the processing time for the 5 trials was averaged. The processing time for object 3 is longer than the other objects due to its smaller pixel area in the frame.

Figure 7 shows the normalized scale (i.e. scaled from 0 to 1) of processing times of all objects with various search windows sizes. It can be seen that in general, for smaller search windows, the processing time was significantly shortened and it gradually approached the processing time for searching the whole window as the search window size increased. This is expected as the search window size approaches the size of the entire video frame. From Figure 7, the normalized scale of average processing time with respect to the different search window sizes can be approximated generally using a polynomial function. In this case, a fifth orders polynomial function is used to relate the search window size 'x' to average time processing 'F(x)' and can be expressed in Eq. (12).

Table 2.
Average processing times (in seconds)

| Window Multiple | Object 1 | Object 2 | Object 3 |
|---|---|---|---|
| 0.5 | 56.78 | 56.68 | 70.36 |
| 1.0 | 59.63 | 58.50 | 68.99 |
| 1.5 | 60.41 | 60.22 | 70.70 |
| 2.0 | 61.58 | 65.55 | 71.23 |
| 2.5 | 61.72 | 68.36 | 73.25 |
| 3.0 | 64.06 | 67.84 | 71.99 |
| 3.5 | 65.13 | 66.92 | 74.22 |
| 4.0 | 65.53 | 66.98 | 75.43 |
| 4.5 | 67.51 | 68.08 | 75.64 |
| 5.0 | 69.26 | 68.80 | 77.40 |
| 5.5 | 69.31 | 69.23 | 78.21 |
| 6.0 | 70.01 | 68.33 | 80.87 |
| 6.5 | 70.07 | 68.76 | 82.04 |
| 7.0 | 69.87 | 67.86 | 82.76 |
| 7.5 | 68.95 | 68.30 | 83.75 |
| 8.0 | 69.19 | 68.81 | 83.32 |
| 8.5 | 68.20 | 67.45 | 81.80 |
| 9.0 | 69.42 | 67.24 | 82.67 |
| 9.5 | 68.83 | 66.84 | 82.91 |
| Full Window | 69.77 | 71.70 | 88.88 |

$$F(x) = a_5 x^5 + a_4 x^4 + a_3 x^3 + a_2 x^2 + a_1 + a_0 \quad (12)$$

Using polynomial curve fitting, "polyfit", function in MATLAB 2014b [14], the approximated coefficients for Eq. (12) can be seen in the Table 3. When the whole frame was searched, the center of the object was accurately found. When the object was detected in a scene by a smaller search window, the detected location was not always the actual center of the object. This was due to the search window not always encompassing the entire object such as when the search window multiple was less than one or when the object was only partly captured in the search window. Figure 8, Figure 9, and Figure 10 illustrate the tracked object centers for object 1, 2 and 3, respectively, for various search window sizes. It can be seen that the detected location converges to the actual center location of the object as the window multiple increases.

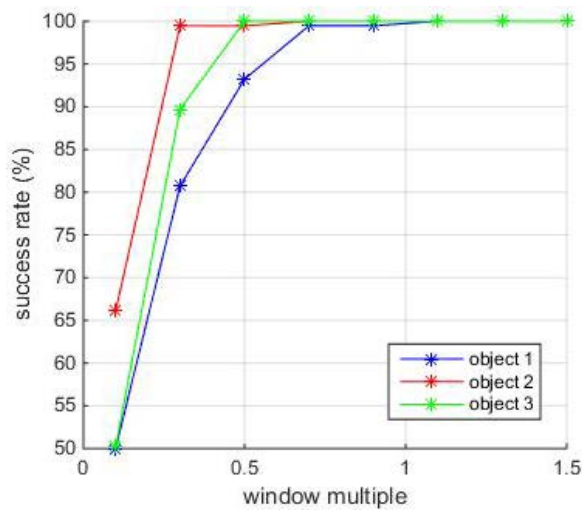

Figure 6. Success rate for tracking objects using various window sizes

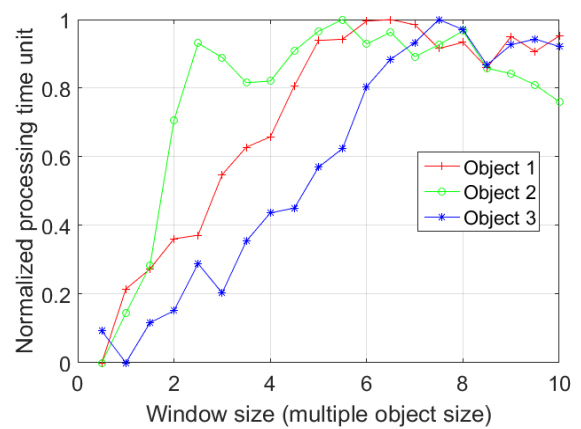

Figure 7. Normalized scale of processing time for tracking objects using various window sizes

Table 3.
Approximated coefficients for polynomial function in equation 12

| Coefficient | Approximated coefficient |
|---|---|
| $a_0$ | -0.2007 |
| $a_1$ | 0.4672 |
| $a_2$ | -0.1370 |
| $a_3$ | 0.0295 |
| $a_4$ | -0.0032 |
| $a_5$ | 0.0001 |

Table 4.
Mean distance error (in pixels)

| Window Multiple | Object 1 | Object 2 | Object 3 |
|---|---|---|---|
| 0.5 | 31.68 | 68.75 | 15.31 |
| 1.0 | 15.83 | 74.11 | 10.86 |
| 1.5 | 10.71 | 78.70 | 9.25 |
| 2.0 | 8.92 | 69.68 | 8.56 |
| 2.5 | 7.73 | 25.58 | 7.48 |
| 3.0 | 6.45 | 0.27 | 7.23 |
| 3.5 | 4.06 | 0.27 | 5.10 |
| 4.0 | 3.04 | 0.27 | 3.29 |
| 4.5 | 1.48 | 0.27 | 2.55 |
| 5.0 | 0.22 | 0.27 | 1.81 |
| 5.5 | 0.52 | 0.27 | 1.19 |
| 6.0 | 0.52 | 0.27 | 0.64 |
| 6.5 | 0.52 | 0.27 | 0.18 |
| 7.0 | 0.52 | 0.27 | 0 |
| 7.5 | 0.52 | 0.27 | 0 |
| 8.0 | 0.52 | 0.27 | 0.55 |
| 8.5 | 0.52 | 0.27 | 0.55 |
| 9.0 | 0.52 | 0.27 | 0.55 |
| 9.5 | 0.52 | 0.27 | 0.55 |

The error between the actual object centers and the tracked centers for various search window sizes can be calculated as the Euclidean distance 'd' between two centers ($(x_1, y_1)$ and $(x_2, y_2)$). This Euclidean distance can be calculated in Eq. (13) [15].

$$d = \sqrt{(x_2 - x_1)^2 + (y_2 - y_1)^2} \quad (13)$$

The Euclidean distance error of the tracking process on all three objects for various search window sizes can be seen in Figure 11. As expected, for larger search window sizes the distance error was lower. The average error in each search window size can be seen in Table 4.

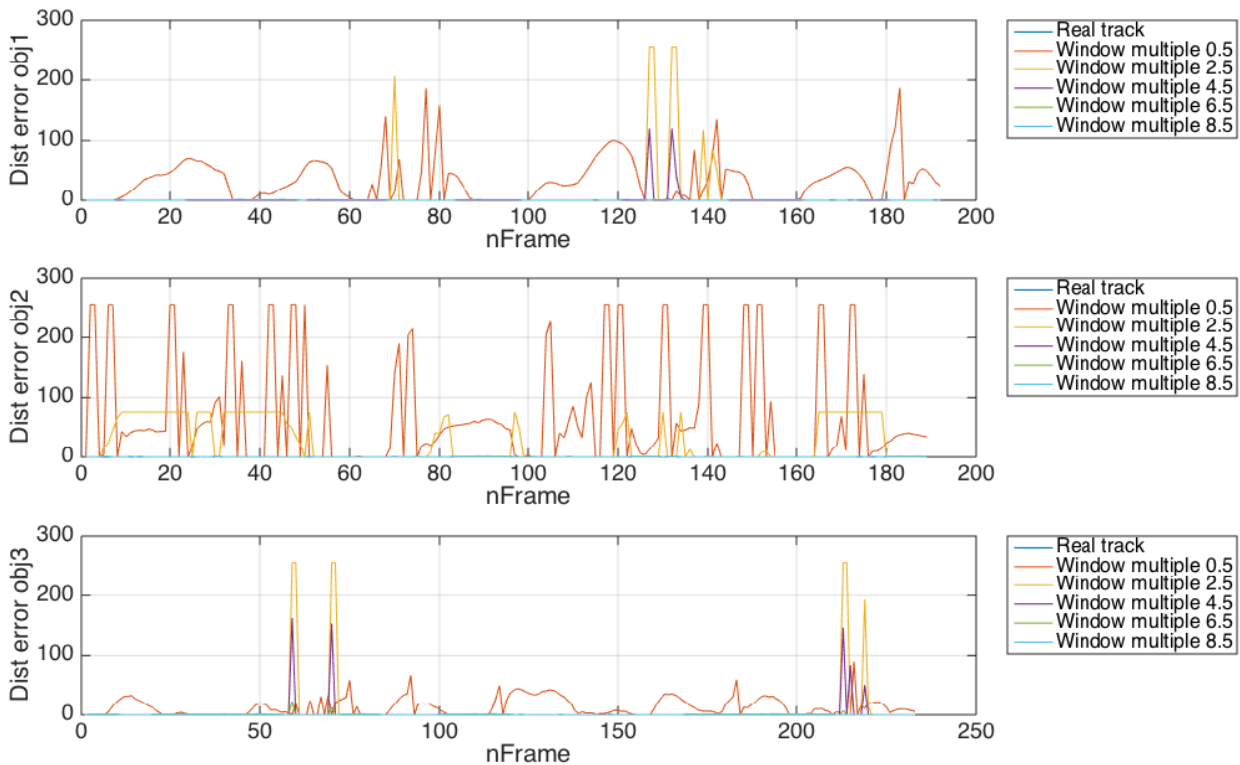

Figure 8. Difference between the detected and predicted center of object 1 and the actual center for various window sizes

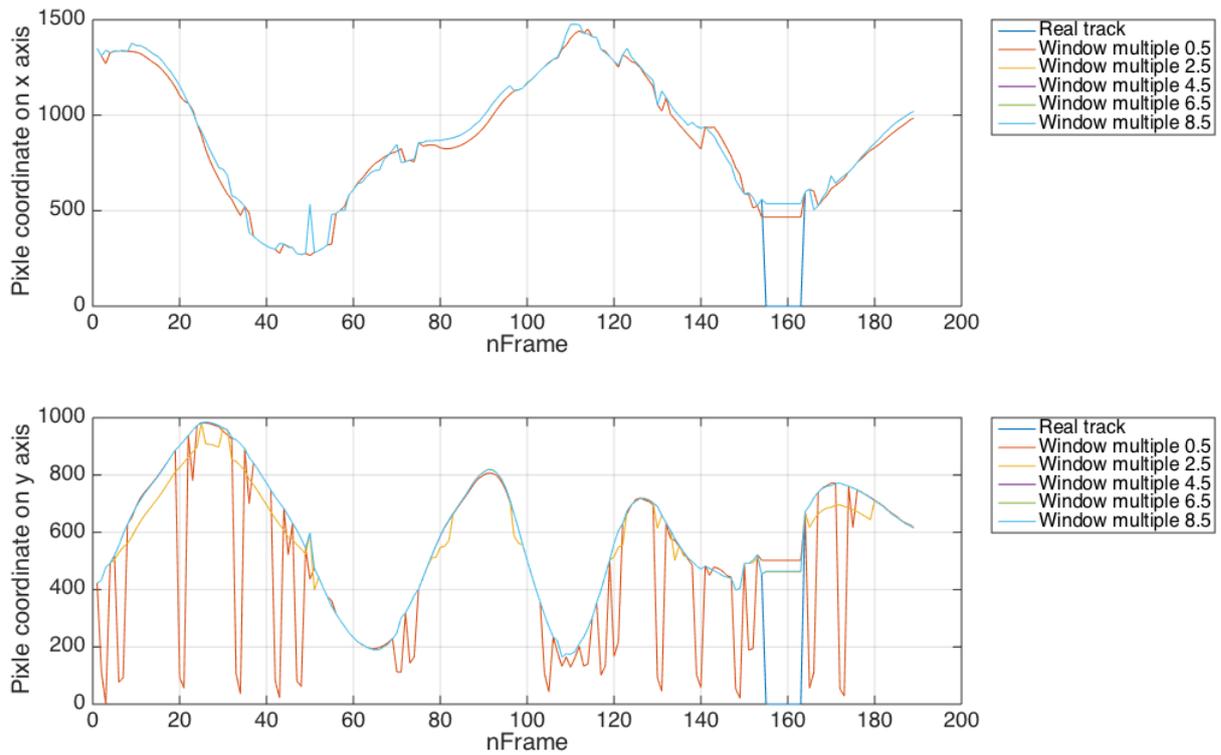

Figure 9. Difference between the detected and predicted center of object 2 and the actual center for various window sizes

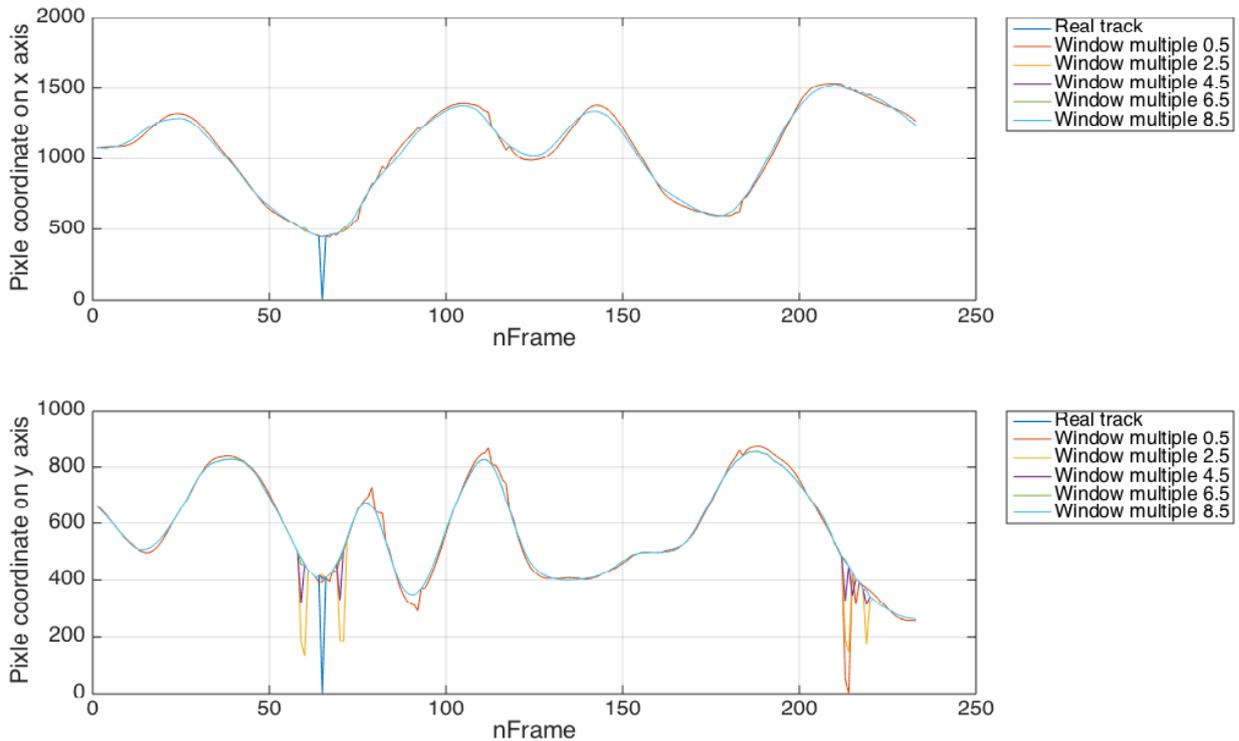

Figure 10. Difference between the detected and predicted center of object 3 and the actual center for various window sizes

Figure 12 shows the normalized scale of the mean distance error for tracking with various search window sizes. From Figure 12, it can be seen that the distance error 'G(x)' with respect to the search window size '$x$' can be approximated using an exponential equation as expressed in Eq. (14).

$$G(x) = ae^{bx} \quad (14)$$

Using the exponential curve fitting function on MATLAB2014b [16], the approximated parameters '$a$' and '$b$' for Eq. (14) can be seen in Table 5.

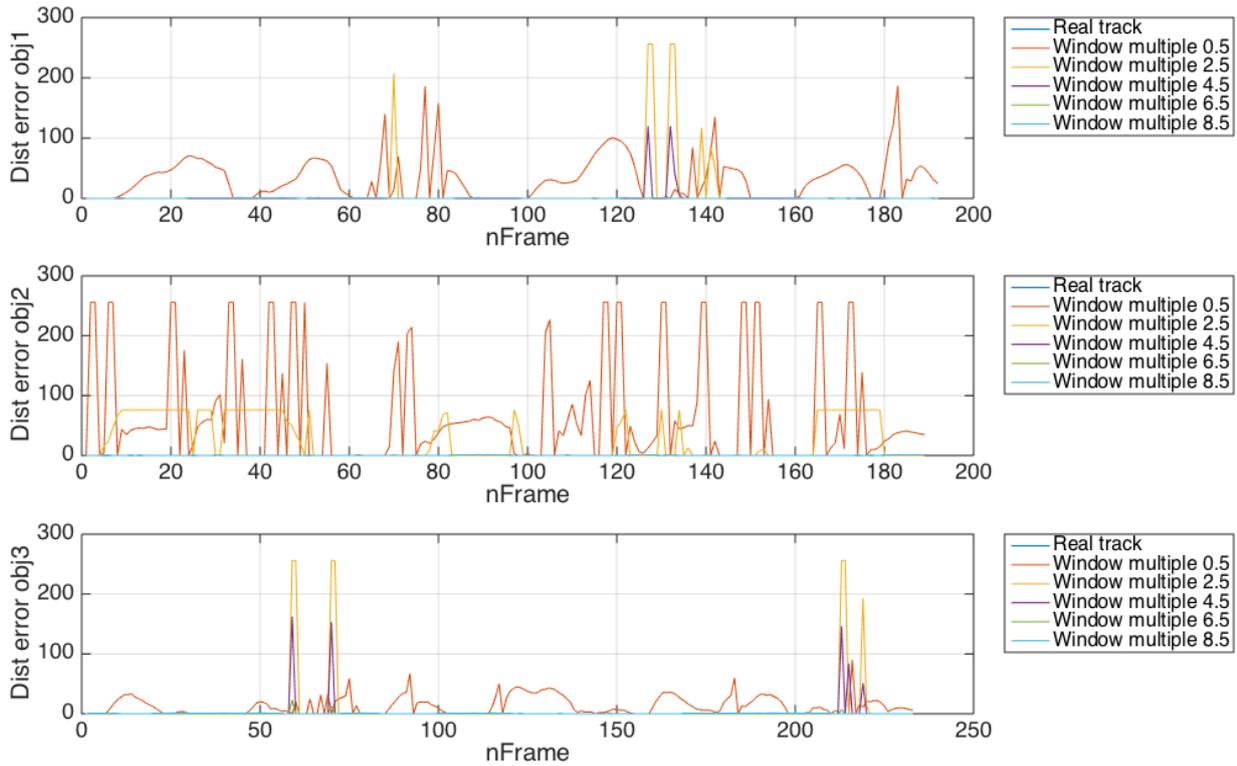

Figure 11. Distance error of the tracking process for various search window sizes

Figure 13 illustrates the relationship between the average processing time and mean distance error with respect to various search window sizes based on the approximated general function F(x) and G(x). To optimize the tracking processes, in this case, it can be done by minimizing the processing time while also minimizing the distance error on the process. The mathematical equation of this optimization process can be expressed in Eq. (15).

$$\arg\min_{x \in [0 \to 10]} \{F(x) + G(x)\} \quad (15)$$

The optimum window size for tracking can be found at the intersection point between the distance error curve and the processing time curve. It is when F(x) and G(x) minimal for a certain window size. Figure 14 shows that these two curves intersect on coordinate (2.16, 0.40). It can be concluded that the optimum window multiple size for this tracking process is 2.16 multiple of the objects largest dimension.

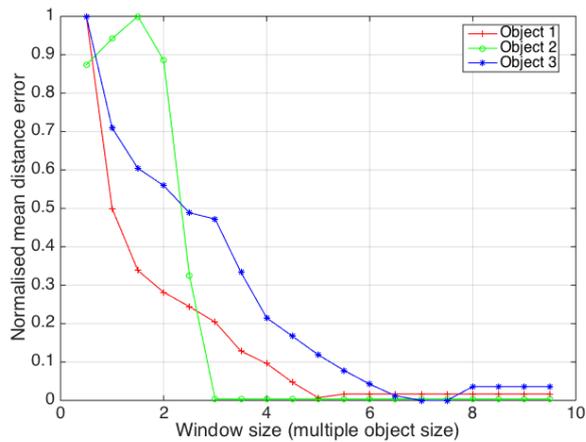

Figure 12. Normalized scale of mean distance error of object tracking using various window sizes

Table 5.
Approximated parameters for equation 14

| Coefficient | Approximated coefficient |
|---|---|
| $a$ | 1.31 |
| $b$ | -0.5457 |

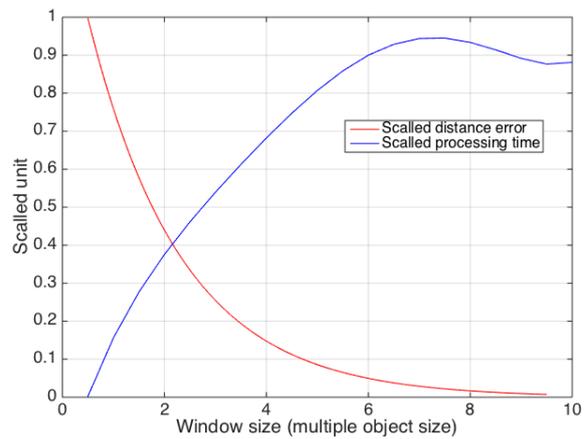

Figure 13. Relation between average time processing and mean distance error in different search window sizes

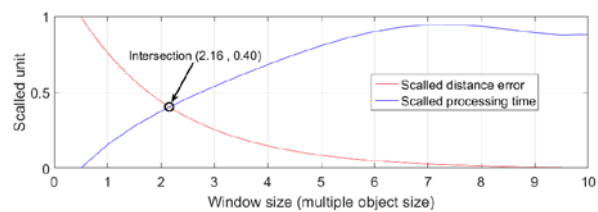

Figure 14. Intersection between distance error curve and time processing curve

## V. CONCLUSIONS

Using a cropped image at a predicted location for an object in a given frame provided a significant improvement in processing time for object tracking compared to conventional methods, which uses whole image size tobe processed, while still maintaining high accuracy in detecting the object in a scene. The processing time improvement was greater when the window size was smaller. The reduction in processing time was also greater for objects with a smaller pixel area. The window multiple of 1 provided the optimum size as it detected all objects 100% of the time and produced a significant improvement in processing time. Certain applications of object tracking may require the center location of the object rather than a location that a part of the object is located in. For these applications, a window multiple larger than 1 for the search window would be required to ensure the search window captured the whole object to determine its center. In this experiment, the process to optimize the search window size choice has been undertaken by minimizing the processing time while also minimizing the distance error. Based on the intersection between distance error curve and time processing curve, it has been obtained that the optimum search window size in this case is about 2.16 times of the object length. Further research could be undertaken in the future to determine the improvement in accuracy and processing time when using a search window for various other object detection algorithms, such as using SURF and SHIFT algorithm.

## ACKNOWLEDGEMENT

The authors would like to thank Australia's New Colombo Plan internship program secured by UNSW for the opportunity and financial support and all those who have helped in conducting this research.